\pdfoutput=1

\documentclass[11pt]{article}

\usepackage[final]{acl}

\usepackage{times}
\usepackage{latexsym}

\usepackage[T1]{fontenc}

\usepackage[utf8]{inputenc}

\usepackage{microtype}

\usepackage{inconsolata}

\usepackage{graphicx}
\usepackage{amsmath}
\usepackage{algorithm}
\usepackage{algpseudocode}

\usepackage{multirow}
\usepackage{array}
\usepackage{url}

\title{JEBS: A Fine-grained Biomedical Lexical Simplification Task}

\author{
 \textbf{William Xia\textsuperscript{1}},
 \textbf{Ishita Unde\textsuperscript{2}},
 \textbf{Brian Ondov\textsuperscript{3}},
 \textbf{Dina Demner-Fushman\textsuperscript{4}}
\\
\\
 \textsuperscript{1}Tufts University,
 \textsuperscript{2}Johns Hopkins University,
 \textsuperscript{3}Yale University,
 \textsuperscript{4}National Library of Medicine,
\\
 \texttt{wxia01@tufts.edu},
 \texttt{iunde1@jhu.edu},
 \texttt{brian.ondov@yale.edu},
 \texttt{ddemner@mail.nih.gov}
}

\begin{document}
\maketitle
\begin{abstract}
Online medical literature has made health information more available than ever, however, the barrier of complex medical jargon prevents the general public from understanding it.
Though parallel and comparable corpora for Biomedical Text Simplification have been introduced, these conflate the many syntactic and lexical operations involved in simplification.
To enable more targeted development and evaluation, we present a fine-grained lexical simplification task and dataset, Jargon Explanations for Biomedical Simplification (JEBS)\footnote{\texttt{https://github.com/bill-from-ri/JEBS-data}}. The JEBS task involves identifying complex terms, classifying how to replace them, and generating replacement text. The JEBS dataset contains 21,595 replacements for 10,314 terms across 400 biomedical abstracts and their manually simplified versions.
Additionally, we provide baseline results for a variety of rule-based and transformer-based systems for the three sub-tasks.
The JEBS task, data, and baseline results pave the way for development and rigorous evaluation of systems for replacing or explaining complex biomedical terms.

\end{abstract}

\section{Introduction}

\begin{figure*}[ht!]
    \centering
    \includegraphics[width=2\columnwidth]{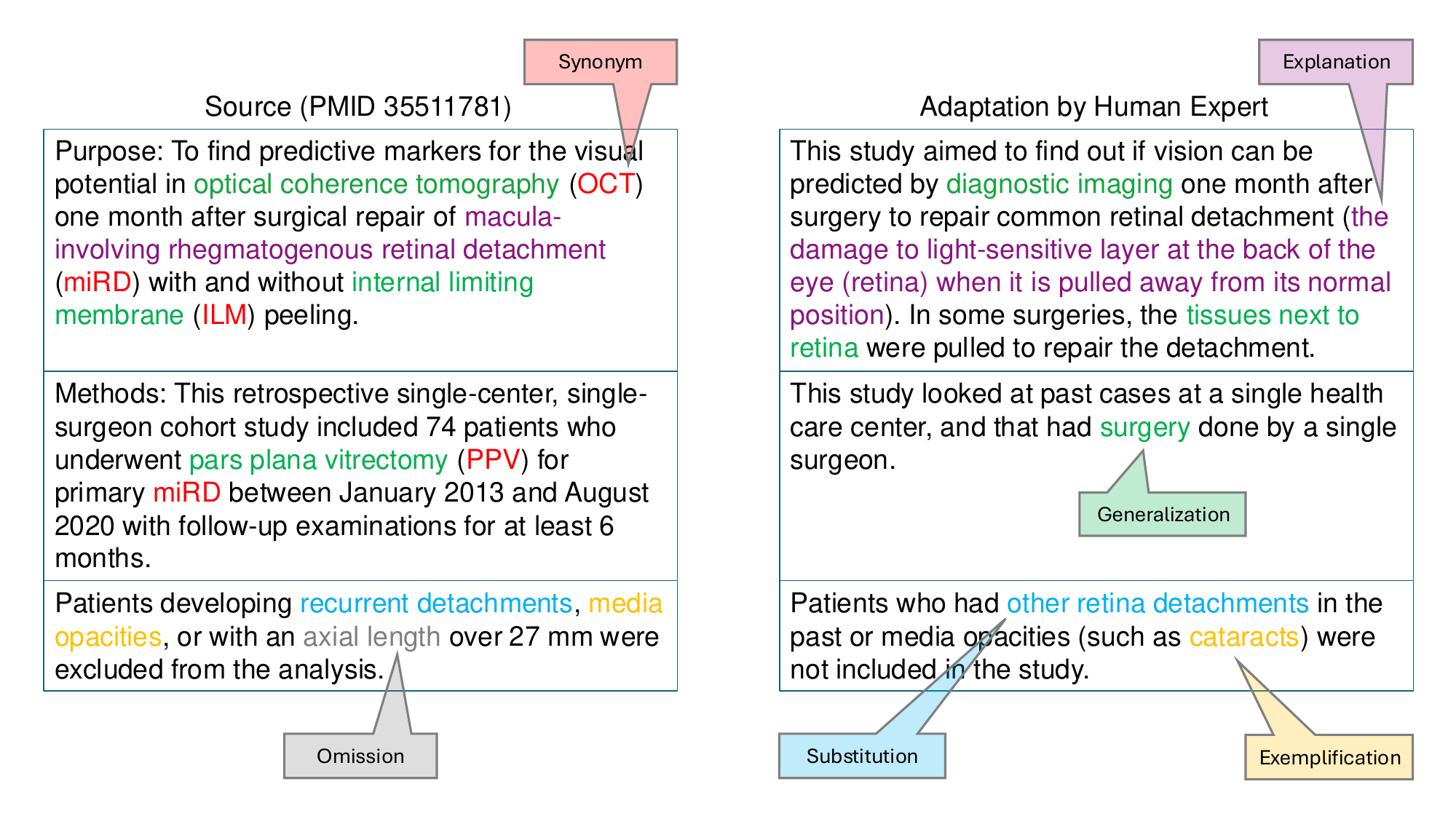}
    \caption{Examples of the JEBS task. Expert terms identified in the source (left) are classified as \textit{substitutions}, \textit{explanations}, \textit{generalizations}, \textit{exemplifications}, or \textit{omissions}. For all types but omissions, the corresponding span in the human expert adaptation is identified (right). Additionally, synonyms (left, red) are identified and linked to the first mention of a term within a synonymous set.}
    \label{adaptations}
\end{figure*}

Understanding medical concepts is critical when making informed healthcare decisions~\cite{kindig2004health}. Patients that lack this understanding are at a disadvantage when making health-related choices, which can negatively affect health outcomes~\citep{king2010poor,berkman2011low}.
Websites such as PubMed~\cite{wheeler2007database} make the latest biomedical knowledge available to everyone. However, because this information is not written for a general audience, attempting to read it without the relevant expertise may cause more harm than good~\cite{white2009cyberchondria}.

Manually curated resources, such as MedlinePlus~\cite{miller2000medlineplus} or UpToDate Patient Education~\cite{fox2003uptodate}, aim to rewrite biomedical knowledge for the public, thus providing a consumer-friendly alternative to resources such as PubMed. However, these resources require massive cost and effort to keep updated with the latest research and are limited in the scope of their topics. For example, UpToDate has more than ten times as many articles written for healthcare practitioners than it has in its Patient Education section. 
Advances in artificial intelligence could help solve this bottleneck by automatically 
`translating' the latest medical research into simpler language or by providing real-time explanations as a reading aid. Given the high stakes of the biomedical domain, however, rigorous evaluation of such systems is crucial.

Existing datasets proposed for training and evaluating Biomedical Text Simplification systems take the form of parallel or comparable corpora~\cite{van2019evaluating,cao2020expertise,Devaraj2021,guo2022cells,Attal2023}. By not explicitly providing term replacements, these datasets restrict the development to end-to-end text simplification systems. The lack of explicit term replacements also restricts automatic evaluation to coarse n-gram or similarity-based metrics, which conflate the many distinct types of word- and sentence-level operations involved in text simplification and can thus lead to misleading results~\cite{alva-manchego-etal-2021-un}.

In this work, we take a step toward more targeted training and evaluation of biomedical text simplification by introducing a manually annotated, fine-grained dataset of multiple lexical simplification operations. We first break the task of lexical simplification into three sub-tasks: (1) \textit{identification} of complex terms, (2) \textit{classification} of how best to replace the terms in context, and (3) \textit{generation} of replacements. Further, for the classification sub-task, we review the literature on lexical simplification to create a taxonomy of five term replacement types: \textit{substitution}, \textit{explanation}, \textit{generalization}, \textit{exemplification}, and \textit{omission}. We then manually annotate expert terms and their replacements found in the PLABA parallel corpus~\cite{Attal2023}, which contains PubMed abstracts paired with expert-written, sentence-by-sentence simplifications. This results in a high-quality dataset of 10,314 \textit{in situ} expert terms identified across 400 original abstracts, and a total of 21,595 replacements for these terms found in simplified versions of these abstracts, each labeled with a replacement type. Examples of identified terms and replacements are shown in Figure~\ref{adaptations}.




Finally, we demonstrate that the JEBS dataset can be used to train and evaluate a variety of rule-based and transformer-based systems to serve as baselines for future development. Transformer models explored included encoder-only models (in both fine-tuning and feature extraction settings), encoder-decoder models (in a fine-tuning setting), and instruction-tuned decoder-only models (in one- and three-shot, in-context learning settings).
In summary, our contributions are as follows:
\begin{itemize}
    \item We define a new, fine-grained lexical simplification task for the biomedical domain.
    \item We provide a manually annotated dataset of 21,595 term replacements with labeled replacement types. 
    \item We report performance of a variety of rule-based and transformer-based baseline systems for each sub-task.
\end{itemize}

\section{Background}


\textbf{Biomedical Simplification Corpora.} Previous datasets developed for biomedical text simplification are mainly comparable (paragraph-level) corpora~\citep{Phatak2022, Devaraj2021,guo2022cells} or parallel (sentence-level) corpora~\citep{Attal2023, cao2020expertise,van2019evaluating}.
As specific edit operations are not annotated in these datasets, they can only be used to train and evaluate end-to-end sentence-level or paragraph-level systems.
While this approach has its advantages, end-to-end neural systems have a higher chance of losing important phrases or altering the meaning of entire sentences during the simplification process than term-focus lexical simplification methods~\citep{Ondov2022-wo}. To maximize faithfulness to the original texts, we thus focus on term-level text simplification, wherein individual expert terms are first identified in a text before they are replaced or explained to make the text more readable as a whole. Perhaps most similar to our work is the Med-EASi dataset~\cite{basu2023medeasi}, which similarly annotates deletions, elaborations, and replacements in two parallel biomedical corpora. JEBS improves on this in several ways. First, our dataset is much larger, totaling 21,595 replacements, as opposed to 1,979. Second, our classification sub-task is finer-grained, distinguishing `elaborations' by whether they are \textit{explanations} or \textit{exemplifications}, and distinguishing `replacements' by whether they are \textit{substitutions} or \textit{generalizations}. Third, our dataset comes from annotating a high-quality, manually written parallel corpus, as opposed to automatically extracted sentence or short passage pairs from larger comparable corpora. Finally, our term pairs are situated within the context of entire parallel documents, providing crucial context. This allows system development and evaluation to consider crucial surrounding  information, for example to disambiguate acronyms.

\textbf{Lexical Simplification Methods.} Previous work in text simplification has explored various methods of term-level simplification. A common method involves the substitution of complex terms with simpler language \citep{basu2023medeasi, Zeng2005}. We define two different types of simplification based off of this approach: substitution, where a close synonym is chosen as the replacement, and generalization, where a more general term is chosen instead. 

Another common form of simplification takes the form of explanations, where additional text is added to the original text to explain complex terms \citep{basu2023medeasi, Elhadad2006, liu2021graphinedatasetgraphawareterminology, elaborations}. While some previous methods generate explanations for terms in isolation, our dataset provides explanations specific to the context in which expert terms appear in biomedical texts. 

One final form of simplification seen in the literature is omission, where complex terms that are not fully relevant to a text are removed entirely \citep{basu2023medeasi, dong2019editntsneuralprogrammerinterpretermodel}. A drawback of previous methods is that their training data for the omission task included simplifications that used different forms of simplification, including adding words and replacing chunks of the original text. By constructing our dataset for simplification at the term level, we hope to isolate omission simplifications for more focused training.

\textbf{Language Models.} The simplest approaches to term-level simplification in the past involved rule-based systems that rely on plain language thesauri and knowledge bases such as the United Medical Language System (UMLS) \citep{lindberg1993unified} to substitute expert terms with lay language \citep{Kandula2010-db}. While such systems demonstrate promising results, they struggle to capture the nuances of grammar, context, and ambiguity that human simplification is able to achieve \citep{Attal2023}. For that reason, most recent work within this domain utilizes deep learning methods, which have seen an explosion of development both within and beyond the realm of text simplification \citep{nisioi-etal-2017-exploring}. In this paper, we evaluate the performance of both rules-based models and neural approaches on our newly-defined biomedical text simplification task, with the intention of exploring the full breadth of text simplification methods to establish definitive benchmarks for our task. 

\section{Task Definition}
\subsection{Task Overview}
The JEBS task is broken into three sub-tasks:

\begin{enumerate}
    \item \textbf{Identification.}
As the first stage in the term simplification process, this sub-task involves labeling terms in a given text as expert terms.
\item \textbf{Simplification Classification.}
Following the identification step of our simplification task, terms are classified by which method should be used to simplify it. 
\item \textbf{Simplification Generation}. Once the simplification type is identified, appropriate text can be generated to replace or clarify the term.
\end{enumerate}

\subsection{Simplification Types}

Below, we explain each type of simplification. 

\subsubsection{Substitution}
Models performing the substitution task generate a close synonym for the term of interest, which replaces the original term in the sentence where it is used. The goal is to simplify the text while retaining its original meaning and brevity.

\subsubsection{Explanation}
Explanatory text (i.e. a definition) is enclosed by parentheses and inserted into the original text immediately following the explained term. 
A clear definition allows the original term to be retained for later use at the cost of brevity. This simplification method is most useful when there isn’t an exact substitution for a term, but there does exist a concise definition for that term. 

\subsubsection{Generalization}
Similar to substitution, the original term is replaced. This method differs from substitution by purposely attempting to subtract unnecessary information from the original term to make the full text more readable to lay consumers.

The use of generalization is heavily context-dependent. As seen in Figure \ref{adaptations}, human annotators may generalize to avoid long, distracting explanations of terms that are not central to the abstract, in this case deeming it only necessary to get across that ``pars plana vitrectomy (PPV)'' was ``surgery.'' However, if a study were comparing PPV to another surgical technique, generalization would not be appropriate. 

\subsubsection{Exemplification}
For exemplification, models generate a short list of examples of the provided expert term. These examples are then inserted into the original text after the term of interest in the same way as in the explanation method. Exemplification can be useful when examples can convey a concept better than a synonym (which may or may not exist) or an definition (which may be long and/or complicated). However, for expert terms that are too specific and therefore lack useful examples, exemplification may not be appropriate. Likewise, if the examples are too esoteric, the text may become more complex, e.g., if the term ``nitrosoureas'' is exemplified with ``carmustine, lomustine, and streptozotocin''. 

\subsubsection{Omission}
The term of interest is removed from its sentence and the original sentence is restructured to make sense without it. For example, removing the entire clause in which the original term occurred. Like generalization, omission seeks to remove unnecessary information from the original text, ideally reducing confusion for consumer readers. However, it may cause passages to become more confusing if too much information is removed or if terms are omitted in ways that leave the sentence grammatically incorrect.

\section{Dataset Creation}

\begin{figure*}
    \centering
    \includegraphics[width=0.9\textwidth]{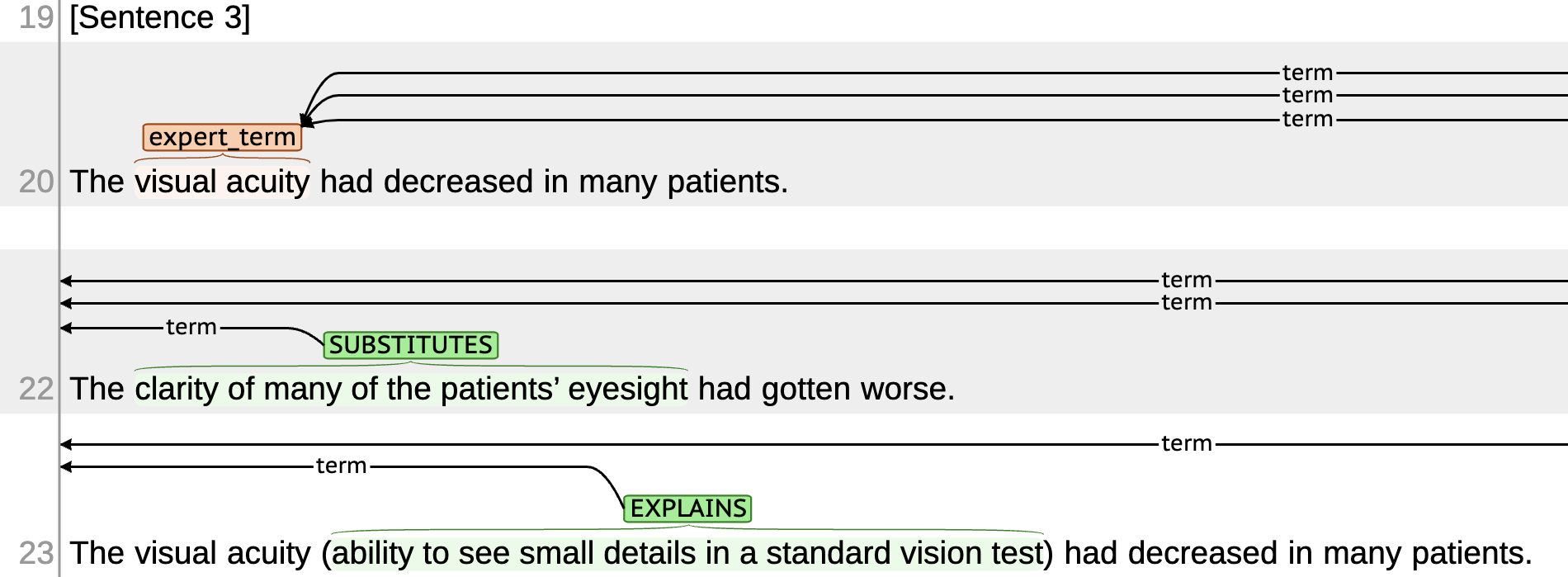}
    \caption{An example annotation of the PLABA dataset, as seen on brat. Line 20 is the original sentence from an abstract; Lines 22 and 23 are from two PLABA simplifications. In each simplification, a replacement span has been identified, in one case being labeled as a substitution, and in the other being labeled as an explanation.}
    \label{brat}
\end{figure*}

The JEBS dataset is derived from 400 abstracts and their associated adaptations
, as found in the PLABA dataset \cite{Attal2023}. Abstracts were aligned at the sentence level with their corresponding adaptations, then annotated by two authors using the brat rapid annotation tool\footnote{\url{http://brat.nlplab.org}} \cite{brat}, which involved selecting expert terms and linking them with their respective simplifications, as found in the PLABA adaptations. In total, the JEBS dataset contains 10,314 expert terms (25.79 terms per abstract) and 21,595 simplifications. Table \ref{counts} displays counts of each simplification type. 

11.47\% of all expert terms in the data appeared alongside acronyms or other names. In the JEBS dataset, expert terms are linked to the simplifications associated with their synonyms. Appendix \ref{synonyms} describes how this linking was performed.

The annotations exhibit a moderate inter-annotator agreement for both the identification task (0.5203 F1) and the classification task (0.4577 F1). Figure \ref{brat} shows an example of the brat interface during annotation. 

\begin{table}[b]
    \centering
    \begin{tabular}{ccc}
        \hline
        \textbf{Simplification Type} & \textbf{Count} & \textbf{Proportion} \\
        \hline
        Substitutions       & 13,966 & 0.6467 \\
        Explanations        & 4,161  & 0.1927 \\
        Omissions           & 1,963  & 0.0909 \\
        Generalizations     & 1,368  & 0.0633 \\
        Exemplifications    & 137    & 0.0063 \\
        \hline
    \end{tabular}
    \caption{Count and Proportion of Simplification Types}
    \label{counts}
\end{table}

While performing annotations, the annotators confirmed that there was no information naming or uniquely identifying individual persons in the PLABA dataset, nor was there any offensive content. The JEBS dataset therefore does not include any such information.

\section{Baseline Systems}
Expert term identification, term classification, and the five forms of simplification were divided into separate sets of language models. All fine-tuned transformer approaches used Hugging Face and PyTorch for fine-tuning. Each of those models underwent 3 epochs of fine-tuning. All fine-tuning and evaluations of those models was performed on a single NVIDIA A100 80GB GPU. 

Prior to training our baseline models, the union of both annotator's annotations were preprocessed into a JSON file, where each expert term in each abstract was linked with its associated simplifications. Each simplification takes the form of a tuple storing both its type (substitution, explanation, generalization, etc) and the contents of that simplification.
All data was split into train and evaluation sets according to a 1:3 ratio. The split was performed at the question-level. That is, data from abstracts answering the same question within the PLABA dataset were kept together in either the train or evaluation sets. Furthermore, neural models designed for the non-identification sub-tasks require the context in which terms were used to function. This data was obtained by splitting PLABA abstracts into individual sentences. 

In the following subsections, we summarize the baseline models for each sub-task, as well as the data preprocessing requirements for each model.

\subsection{Identification}
The rule-based identifier model uses MetaMapLite \cite{metamaplite}, the Unified Medical Language System (UMLS) \cite{lindberg1993unified}, and two term frequency datasets from Kaggle---one derived from the Google Web Trillion Word Corpus \cite{unigrams} and the other derived from BookCorpus and a 2019 dump of Wikipedia \cite{bigrams}---to identify and filter expert terms.

After the rule-based model, we fine-tuned a set of transformer-based identifier models using pretrained versions of BERT Large (340M parameters) \cite{bert}, BioBERT Large (340M parameters) \cite{biobert}, XLM RoBERTa Large (550M parameters) \cite{xlm-roberta}, and DeBERTa Large (435M parameters) \cite{deberta}. For these models, we framed the sub-task as a named entity recognition (NER) problem \citep{Bose2021}. Abstracts were therefore preprocessed for the fine-tuned identifier models by labeling each sentence according to a Beginning-Inside-Outside labeling scheme. Because the goal of this sub-task was purely to identify expert terms, labeling was performed without consideration for the simplification types that could be assigned to each term.

In addition, we evaluated Llama3 Instruct's (8B parameters) \cite{dubey2024llama} performance on this task, instructing the LLM to perform the identification task on a single sentence. The sentence to operate on was provided immediately after the following instruction prompt was provided:
\begin{quote}
    \textbf{Prompt:} ``Identify all non-consumer biomedical terms in the user's sentence using a comma-separated list. Generate no other text besides the list.''
\end{quote}


Four different metrics were used to evaluate the identification models. The first was the average F1 score, which was computed for a given model by finding its F1 score against both annotator's individual annotations, then averaging the results.
Union and intersection F1 scores were taken according to the union and intersection of the two annotators' identified terms. Finally, models were evaluated according to a Pyramid score \cite{pyramid}, where points were given for each expert term depending on how many annotators identified it as an expert term, then normalized according to the maximum score each model could have attained.

Running the rule-based model on the JEBS test data set for evaluation took 8 minutes and 34 seconds to run on an Apple M1. Training the BERT-based transformer models on the training data took around 2 minutes and 58 seconds each. Running those models for evaluation on the test data took around 2 minutes and 26 seconds each. Finally, Llama3 took 28 minutes and 57 seconds for the identification task on the JEBS test data.

\subsection{Simplification Classification}
For the classification task, we divided our approaches into a frozen-weights transformer-based method and a fine-tuned transformer-based method. 

In the former, we preprocessed abstract sentences by indicating the expert terms within them using custom tokens \texttt{<ext>} and \texttt{</ext>}. Preprocessed sentences were embedded using BERT-Large and DeBERTa-Large before being used to train and evaluate two separate multi-label classifier models. These classifiers were built using PyTorch neural networks. The BERT and DeBERTa multi-label classifier models took 20 seconds and 24 seconds respectively to run on the test data.

In the second approach, we combined the identification and classification sub-tasks by framing classification as a slightly more advanced NER problem. The data for this approach took the form of BIO-labeled sentences, where terms were labeled with the simplification method assigned to them most often in the training data. 
Pretrained versions of BERT-Large and DeBERTa-Large were fine-tuned using the preprocessed data to distinguish between non-expert terms and terms that should be simplified using one of each simplification methods described in this paper. These models therefore performed both the identification and classification sub-tasks at the same time. The BERT and DeBERTa NER models took 36 seconds and 89 seconds respectively to run on the test data.

Outputs were evaluated according to two metrics: average F1 score and union F1 score. These metrics were computed according to the labels assigned to expert terms by both annotators separately, and the union of labels assigned to expert terms by both annotators, respectively. Scores were macro-averaged across the five simplification methods to account for the class imbalance in our data.

\subsection{Simplification Generation}\label{1c_explanation}

We evaluated the performance of three off-the-shelf language models on each simplification method: Llama3-8B Instruct, GPT-4o-mini \cite{openai2024gpt4ocard}, and Llama4 Scout Instruct \cite{llama4}. For all simplification methods, the input sequence took the form of a sentence with a single expert term highlighted via enclosing brackets. Sentences containing multiple expert terms were duplicated in our data with a different expert term selected. With the exception of omission, output sequences for all simplification methods were composed entirely of the generated simplification. 

In addition to the simplification instruction, we provided each language model with either 1 or 3 example simplifications to leverage in-context learning \citep{brown2020languagemodelsfewshotlearners}. Prompting is described in greater detail in Appendix~\ref{app:prompts}.

Because our goal for these baselines was to gauge out-of-the-box performance, rather than develop the best possible models, we did not fine-tune LLMs or perform rigorous hyperparameter tuning. We therefore set the main hyperparameters within common ranges:
\begin{itemize}
    \setlength\itemsep{-0.5em}
    \item \texttt{max\_new\_tokens} = 512
    \item \texttt{do\_sample} = True
    \item \texttt{temperature} = 0.6
    \item \texttt{top\_p} = 0.9
\end{itemize}
Running Llama3 on the JEBS test set took around 17 minutes for each simplification method.

Evaluations were performed manually on subsets of 48 randomly chosen replacements for each of the 5 simplification types and for 6 baselines, along 4 axes (simplicity, accuracy, completeness, and brevity), totaling $48*5*6*4=5,760$ total judgments. Each axis was rated by authors (all biomedical informatics experts) on a 5-point symmetric likert scale, which was interpolated to a 0-1 score for reporting. Below is an explanation of the 4 metrics used to evaluate model outputs for the generation sub-task.
\begin{itemize}
    \item \textbf{Simplicity:} The simplification is easy to understand.
    \item \textbf{Accuracy:} The simplification contains accurate information.
    \item \textbf{Completeness:} The simplification minimizes information lost from the original text.
    \item \textbf{Brevity:} The simplification is concise.
\end{itemize}


\subsection{End-to-End System}
We combined the DeBERTa identification baseline, BERT NER classification baseline, and 3-shot prompted Llama 3 generation baseline to create an end-to-end text simplification system. By running this system against our test data, we simulated the real-world use case of our baselines and investigated potential errors that may propagate through our simplification pipeline.

Because the BERT NER classification baseline performs identification as well as classification, it can classify terms not selected by the DeBERTa identification baseline. Therefore, to reduce false positives when identifying expert terms, we only performed classification on terms identified by both the identification and classification baselines. Because this was the only part of the end-to-end system where the components did not work independently of one another, our analysis  was focused on identification.

\section{Results}
\begin{table*}
    \centering
    \begin{tabular}{|c|c|c|}
        \hline
        \textbf{Input} & \textbf{Model} & \textbf{Identified Terms} \\
        \hline
        \multirow{7}{12em}{“Ring sutures induced cataract more frequently than other procedures.”} 
        & Gold Standard & \multirow{1}{15.4em}{Ring sutures, cataract} \\
        \cline{2-3}
        & Rule-based & \multirow{1}{15.4em}{sutures, cataract} \\
        \cline{2-3}
        & BERT-L & \multirow{1}{15.4em}{Ring sutures, cataract} \\
        \cline{2-3}
        & BioBERT-L & \multirow{1}{15.4em}{Ring sutures, cataract} \\
        \cline{2-3}
        & XLM RoBERTa-L & \multirow{1}{15.4em}{Ring sutures, cataract} \\
        \cline{2-3}
        & DeBERTa & \multirow{1}{15.4em}{Ring, cataract} \\
        \cline{2-3}
        & Llama3 & \multirow{1}{15.4em}{sutures, cataract} \\
        \hline
    \end{tabular}
    \caption{\label{id_example}Example input sentence and terms identified by each identifier model.}
\end{table*}

\subsection{Identification}
The transformer-based models outperformed the rule-based model in the identification sub-task, with the DeBERTa-based model achieving the highest score in all four metrics. Interestingly, despite being pretrained on domain knowledge, BioBERT fails to outperform the BERT-based identification model. It seems that in the identification sub-task, domain knowledge doesn't enhance LLM performance.


\begin{table*}
    \centering
    \begin{tabular}{ccccc}
        \hline
        \textbf{Model} & \textbf{Avg F1} & \textbf{$\cup$ F1} & \textbf{$\cap$ F1} & \textbf{Pyramid} \\
        \hline 
        Rule-based & 0.2097 & 0.2487 & 0.1497 & 0.2916 \\
        BERT-L & 0.3530 & 0.4260 & 0.2515 & 0.4891 \\
        BioBERT-L & 0.3058 & 0.3898 & 0.2071 & 0.3938 \\
        XLM RoBERTa-L & 0.3745 & 0.4596 & 0.2578 & 0.5147 \\
        DeBERTa-L & \textbf{0.4317} & \textbf{0.5255} & 0.2976 & 0.6014 \\
        Llama3 & 0.3678 & 0.4085 & 0.3095 & 0.4692 \\
        \hline
        $\text{BERT-L}_{cls}$ & 0.2785 & 0.3399 & 0.1955 & 0.3895 \\
        $\text{DeBERTa-L}_{cls}$ & 0.3448 & 0.4009 & 0.2628 & 0.4564 \\
        \hline
        $\text{End-to-end}$ & 0.3923 & 0.3918 & \textbf{0.4304} & \textbf{0.6207} \\
        \hline
    \end{tabular}
    \caption{\label{id_performance}Performance of each identifier model, the NER classification models, and the end-to-end system.}
\end{table*}

\subsection{Simplification Classification}
Among the frozen-weights transformer approaches, the classifier trained on DeBERTa sentence embeddings performed better during evaluation, though neither model was especially effective at classifying expert terms.

The NER models outperformed the neural networks used for this task. However, their ability to perform classification came at the cost of lowered overall term identification accuracy. Compared to the identification models fine-tuned on the same base models, the NER models fine-tuned for this task under-performed when identifying expert terms. The performance of the NER-based models can be found in Table \ref{id_performance}.

\begin{table}
    \centering
    \begin{tabular}{ccc}
        \hline
        \textbf{Model} & \textbf{Avg F1} & \textbf{$\cup$ F1}  \\
        \hline
        BERT Frozen & 0.0337 & 0.0334 \\
        DeBERTa Frozen & 0.1823 & 0.1856 \\
        BERT NER & \textbf{0.3588} & \textbf{0.3413} \\
        DeBERTa NER & 0.3300 & 0.3363 \\
        \hline
    \end{tabular}
    \caption{\label{classification}Results on the simplification classification task.}
\end{table}

\subsection{Simplification Generation}

\begin{table*}
    \centering
    \begin{tabular}{cccccc}
        \hline
        \textbf{Model} & \textbf{SUB} & \textbf{EXP} & \textbf{GEN} 
        & \textbf{EXE} & \textbf{OMI}\\
        \hline
        Llama3, 1-shot      & 0.7878 & 0.4648 & 0.8398 & 0.7396 & 0.7005 \\
        Llama3, 3-shot      & 0.8529 & 0.4922 & 0.8045 & 0.7669 & 0.7474 \\
        Llama4, 1-shot      & 0.7643 & 0.8789 & \textbf{0.9271} & 0.5482 & 0.6966 \\
        Llama4, 3-shot      & 0.8008 & \textbf{0.9180} & 0.9063 & 0.5143 & 0.6953 \\
        GPT-4o, 1-shot      & 0.8333 & 0.5143 & 0.9036 & 0.8008 & 0.7448 \\
        GPT-4o, 3-shot      & \textbf{0.9219} & 0.5443 & 0.8551 & \textbf{0.8216} & \textbf{0.7539} \\
        \hline
    \end{tabular}
    \caption{\label{simp_performance} Aggregate evaluation results of each baseline model on each simplification method for the generation sub-task. Appendix \ref{app:1c_results} shows results for each metric.}
\end{table*}

The new human evaluation results show that the 3-shot prompted GPT-4o-mini baseline generally outperformed the others. Additionally, the new results show that our baselines are best at substitution and generalization, while usually underperforming in explanation and omission. They reveal a clear tradeoff between completeness and brevity, where high performance in one metric typically comes with lower performance in the other. Further error analysis shows that a major source of low `accuracy' ratings was in fact due to the models explaining the wrong term in the input sentence. This provides a clear future direction for prompt engineering to further improve results.

Below, we further detail the performance of our baseline models for each of the five simplification methods.

\subsubsection{Substitution}
For the substitution task, our models consistently generated helpful synonyms for expert terms in our dataset. However, they tended to underperform in terms of completeness, suggesting that information is easily lost even when models successfully generate accurate synonyms. 

\subsubsection{Explanation}
The primary limitation of explanations is that the generated definition can introduce new jargon that complicates the text rather than clarifies it. This is evident from our models performing worst on the simplicity metric for the explanation task, suggesting that the reading level of the generated text was too high. 

\subsubsection{Generalization}
As with the substitution sub-task, models performing generalization did worst on the completeness metric, further showing that even synonyms tend to lose information. However, they did so with a large gain in simplicity, showing that using more general language from the text improves readability. 

\subsubsection{Exemplification}
In most cases, our models generated valid examples for terms tagged for this form of simplification. However, much like with explanation, generated examples often further complicated the text rather than clarifying it, as seen in the low simplicity scores for models attempting this task. This suggests that exemplification should only be used on terms with commonly-known examples.

\subsubsection{Omission}
Omission is a particularly challenging task, as it requires the model to reshape the entire sentence (as opposed to a single term) to make sense following the omission of the term of interest. Our baselines often rewrote the original sentence with the expert term replaced with a synonym instead of removing it entirely, thereby performing substitution instead of omission.

\subsection{End-to-End System}
The combination of the DeBERTa identification model and the BERT classification model yields a strong baseline for the identification task, outperforming all others on the intersection F1 and Pyramid score metrics. The end-to-end system also scores second-highest on the average F1 metric, while performing below average on the union F1 metric. 

These results indicate that cross-referencing terms identified by two models improves precision in the identification sub-task at the cost of recall. 

\section{Future Work}
There remains ample space for improving performance in all of the sub-tasks and methods defined in this paper. For example, it remains to be seen if decoder-only LLMs can effectively perform the identification. While Llama3-8B was unable to outperform most of the encoder-based models, more specific prompt engineering may unlock greater levels of performance. 

In the simplification classification sub-task, there exist multiple unexplored directions from which one could improve upon our baselines. For example, this task could be framed as a sequence-to-sequence problem for generative models to attempt. The issue of class imbalance in the data for this task (wherein the majority of expert terms can be simplified using substitution) must also be addressed, whether that be via class weights, oversampling, or using generative AI to synthesize additional example data.

While off-the-shelf language models performed well on the substitution, explanation, generalization, and exemplification simplification methods, they do not make full use of the JEBS dataset. To fully utilize our dataset, future approaches should explore fine-tuning language models for each simplification method to achieve stronger performance on the generation sub-task.

Finally, the omission task presents a unique challenge in the form of grammar error correction, which we have yet to fully explore. Effective grammar correction with LLMs may be achieved with prompt engineering or fine-tuning on dedicated grammar correction datasets.

\section{Conclusion}
In this work, we introduced a new task of fine-grained biomedical lexical simplification and a corresponding dataset called JEBS (Jargon Explanations for Biomedical Simplification). The JEBS task involves identifying expert terms, classifying how best to replace them, and generating replacement text. Unlike existing parallel or comparable corpora for Biomedical Text Simplification, JEBS allows targeted development and evaluation of systems to directly provide replacement terms. The JEBS dataset contains 21,595 replacements for 10,314 terms. These terms appear in the context of 400 biomedical abstracts and their corresponding manually written plain language adaptations from the PLABA dataset. 
We have introduced a suite of baseline models for identifying expert terms in biomedical texts, classifying them for simplification, and generating consumer-friendly simplifications for those terms. Using an array of methods built atop the JEBS dataset, we achieved promising results in all of our defined tasks. Finally, we proposed avenues for future improvement of our models. We imagine that our work will bridge the gap between medical experts and patients, providing consumers with new tools to aid in healthcare decision making. 

\section{Limitations}
Within the JEBS dataset, there exists a class imbalance between the five simplification types, with substitutions making up a disproportionately large percentage of the total simplifications. This imbalance may limit the effectiveness of future models fine-tuned for classifying terms as well as for generating text for the less common simplification types. Exemplification is especially challenging to fine-tune on, less than 1 percent of the simplifications in the JEBS dataset are exemplifications. 

Although our manual evaluations provide a more accurate and nuanced measure of text simplification than automated metrics, more rigorous methods could yield more reliable results. For example, having multiple experts review each output could reduce the subjectivity associated with evaluating text simplification models. Additionally, performing manual evaluations of a larger subset of the data could provide more accurate insights into overall model performance.

\section{Acknowledgements}
This research was supported by the Division of Intramural Research (DIR) of the National Library of Medicine (NLM), National Institutes of Health.
This work utilized the computational resources of the NIH HPC Biowulf cluster\footnote{https://hpc.nih.gov/}.

\bibliography{references}

\appendix



\section{Linking Synonym Terms}\label{synonyms}
During the annotation of the PLABA dataset, the annotators could select terms as synonyms of other terms. We developed Algorithm \ref{alg:syn_link} to merge the simplifications of synonymous terms when processing the annotations for the JEBS dataset. 

The algorithm takes two dictionaries as input: $D$, which maps terms to their simplifications, and $S$, which maps expert terms to their synonyms. The outputs of the algorithm is a new dictionary $D'$, which maps expert terms to their simplifications and the simplifications of their synonyms.

\begin{algorithm}[t]
\caption{Associate Synonyms Algorithm}\label{alg:syn_link}
\begin{algorithmic}[1]
    \Require Dictionary $D$ mapping terms to simplifications, dictionary $S$ mapping terms to synonyms
    \Ensure Dictionary $D'$ with merged synonyms
    \State $D' \gets \emptyset$
    \ForAll{$(t, sns) \in S$}
        \ForAll{$sn \in sns$}
            \If{$sn \neq t$}
                \ForAll{$sms \in D [t]$}
                    \State $D'[sn] \gets D[sn] \cup sms$
                \EndFor
            \EndIf
        \EndFor
    \EndFor
    \State \Return $D'$
\end{algorithmic}
\end{algorithm}

\begin{table*}[htb!]
    \centering
    \begin{tabular}{{p{0.15\linewidth} | p{0.8\linewidth}}}
         \textbf{Generation} & \textbf{Prompt} \\
         \hline
         Substitution &     ``Provide a simpler term to replace the term in square brackets. The term should be understandable by a general audience. Provide only the replacement term and not the entire sentence.'' \\
         \hline
         Explanation & ``Provide a concise, simple explanation of the term in square brackets.  The explanation should be understandable by a general audience and short enough to put in parentheses after the term. Provide only the explanation and not the entire sentence.'' \\
         \hline
         Generalization & ``Provide a more general term to replace the term in square brackets. The new term should be understandable by a general audience. Provide only the replacement term and not the entire sentence.'' \\
         \hline
         Exemplification &     ``Provide one to three brief example terms to help explain the term in square brackets. The example(s) should be understandable by a general audience and short enough to put in parentheses after the term. Provide only the example(s) and not the entire sentence.'' \\
         \hline
         Omission & ``Rewrite the sentence so it avoids using the term square brackets. The new sentence should be understandable by a general audience. Generate no other text.''
    \end{tabular}
    \caption{Prompts provided to Llama 3, GPT-4o, and Llama 4 for the generation sub-task, by replacement type. Each prompt was directly followed by either 1 or 3 examples and the preprocessed source text to operate on.}
    \label{tab:prompts}
\end{table*}

\section{Generation Sub-task Prompts}\label{app:prompts}
Llama3-8B Instruct, GPT-4o-mini, and Llama 4 Scout were provided with unique prompts for each simplification method used for the generation sub-task. Each time a model was tasked with simplifying a given sentence, the instruction prompt was given first, directly followed by 1 to 3 examples (for 1-shot or 3-shot prompting, respectively) and the source text to operate on. The source text was preprocessed such that the term of interest was enclosed by brackets. Complete instruction prompts are shown in Table~\ref{tab:prompts}.

\section{Generation Sub-task Full Results}\label{app:1c_results}
Figures \ref{fig:simplicity}, \ref{fig:accuracy}, \ref{fig:completeness}, \ref{fig:brevity} show the results for the generation sub-task baselines on each of the manual evaluation metrics described in Section \ref{1c_explanation}.

\begin{table*}
    \centering
    \begin{tabular}{cccccc}
        \hline
        \textbf{Model} & \textbf{SUB} & \textbf{EXP} & \textbf{GEN} 
        & \textbf{EXE} & \textbf{OMI}\\
        \hline
        Llama3, 1-shot  & 0.8177 & 0.4167 & 0.7865 & 0.7448 & 0.4844 \\
        Llama3, 3-shot  & 0.8125 & 0.4375 & 0.9043 & 0.9063 & 0.6927 \\
        Llama4, 1-shot  & 0.8073 & 0.9063 & 0.9688 & 0.4688 & 0.5990 \\
        Llama4, 3-shot  & 0.8177 & 0.9271 & 0.9427 & 0.4948 & 0.7031 \\
        GPT-4o, 1-shot  & 0.8177 & 0.4896 & 1.0000 & 0.9531 & 0.6719 \\
        GPT-4o, 3-shot  & 0.8750 & 0.5208 & 0.9787 & 0.9479 & 0.7344 \\
        \hline
    \end{tabular}
    \caption{\label{fig:simplicity} Human evaluation results of simplicity for each baseline model on each simplification method for the generation sub-task.}
\end{table*}

\begin{table*}
    \centering
    \begin{tabular}{cccccc}
        \hline
        \textbf{Model} & \textbf{SUB} & \textbf{EXP} & \textbf{GEN} 
        & \textbf{EXE} & \textbf{OMI}\\
        \hline
        Llama3, 1-shot  & 0.8333 & 0.5260 & 0.8229 & 0.7604 & 1.0000 \\
        Llama3, 3-shot  & 0.9375 & 0.4063 & 0.7394 & 0.9427 & 0.9740 \\
        Llama4, 1-shot  & 0.8333 & 0.9323 & 0.9323 & 0.6146 & 0.9479 \\
        Llama4, 3-shot  & 0.8490 & 0.9844 & 0.9479 & 0.5781 & 0.9167 \\
        GPT-4o, 1-shot  & 0.9427 & 0.3698 & 0.8698 & 0.9844 & 0.9896 \\
        GPT-4o, 3-shot  & 0.9479 & 0.5833 & 0.8191 & 0.8594 & 0.9740 \\
        \hline
    \end{tabular}
    \caption{\label{fig:accuracy} Human evaluation results for accuracy of each baseline model on each simplification method for the generation sub-task.}
\end{table*}

\begin{table*}
    \centering
    \begin{tabular}{cccccc}
        \hline
        \textbf{Model} & \textbf{SUB} & \textbf{EXP} & \textbf{GEN} 
        & \textbf{EXE} & \textbf{OMI}\\
        \hline
        Llama3, 1-shot  & 0.7448 & 0.5208 & 0.7969 & 0.8542 & 0.9740 \\
        Llama3, 3-shot  & 0.8281 & 0.5469 & 0.6330 & 0.9740 & 0.8854 \\
        Llama4, 1-shot  & 0.7031 & 0.8854 & 0.8490 & 0.5781 & 0.8698 \\
        Llama4, 3-shot  & 0.6875 & 0.9427 & 0.7656 & 0.5260 & 0.7344 \\
        GPT-4o, 1-shot  & 0.8281 & 0.5313 & 0.7500 & 0.9427 & 0.9323 \\
        GPT-4o, 3-shot  & 0.8854 & 0.5573 & 0.6489 & 0.8802 & 0.8698 \\
        \hline
    \end{tabular}
    \caption{\label{fig:completeness} Human evaluation results for completeness of each baseline model on each simplification method for the generation sub-task.}
\end{table*}

\begin{table*}
    \centering
    \begin{tabular}{cccccc}
        \hline
        \textbf{Model} & \textbf{SUB} & \textbf{EXP} & \textbf{GEN} 
        & \textbf{EXE} & \textbf{OMI}\\
        \hline
        Llama3, 1-shot  & 0.7552 & 0.3958 & 0.9531 & 0.5990 & 0.3438 \\
        Llama3, 3-shot  & 0.8333 & 0.5781 & 0.9415 & 0.2448 & 0.4375 \\
        Llama4, 1-shot  & 0.7135 & 0.7917 & 0.9583 & 0.5313 & 0.3698 \\
        Llama4, 3-shot  & 0.8489 & 0.8177 & 0.9688 & 0.4583 & 0.4271 \\
        GPT-4o, 1-shot  & 0.7448 & 0.6667 & 0.9948 & 0.3229 & 0.3854 \\
        GPT-4o, 3-shot  & 0.9792 & 0.5156 & 0.9734 & 0.5990 & 0.4375 \\
        \hline
    \end{tabular}
    \caption{\label{fig:brevity} Human evaluation results for brevity of each baseline model on each simplification method for the generation sub-task.}
\end{table*}

\end{document}